\documentclass[letterpaper]{article} % DO NOT CHANGE THIS
\usepackage{aaai24}  % DO NOT CHANGE THIS
\usepackage{times}  % DO NOT CHANGE THIS
\usepackage{helvet}  % DO NOT CHANGE THIS
\usepackage{courier}  % DO NOT CHANGE THIS
\usepackage[hyphens]{url}  % DO NOT CHANGE THIS
\usepackage{graphicx} % DO NOT CHANGE THIS
\usepackage{natbib}  % DO NOT CHANGE THIS AND DO NOT ADD ANY OPTIONS TO IT
\usepackage{caption} % DO NOT CHANGE THIS AND DO NOT ADD ANY OPTIONS TO IT
\usepackage{algorithm}
\usepackage{algorithmic}
\usepackage{adjustbox}

\usepackage{newfloat}
\usepackage{listings}
\DeclareCaptionStyle{ruled}{labelfont=normalfont,labelsep=colon,strut=off} % DO NOT CHANGE THIS
\floatstyle{ruled}
\newfloat{listing}{tb}{lst}{}
\floatname{listing}{Listing}
\usepackage{xcolor}
\usepackage{booktabs}
\usepackage{tabularx}
\usepackage{multirow}

\usepackage{amsthm}
\usepackage{amsmath} % For equation formatting
\usepackage{amssymb} % For \mathbb
\usepackage{lipsum}
\newcommand{\tablefontsize}{\small} % Define a custom font size for the tables
\title{Controllable Game Level Generation: Assessing the Effect of Negative Examples in GAN Models}
\author {
    Mahsa Bazzaz,
    Seth Cooper
}
\affiliations {
    Khoury College of Computer Sciences, Northeastern University\\
    bazzaz.ma@northeastern.edu, se.cooper@northeastern.edu
}

\begin{document}

\maketitle

\begin{abstract}
Generative Adversarial Networks (GANs) are unsupervised models designed to learn and replicate a target distribution. The vanilla versions of these models can be extended to more controllable models. Conditional Generative Adversarial Networks (CGANs) extend vanilla GANs by conditioning both the generator and discriminator on some additional information (labels). Controllable models based on complementary learning, such as Rumi-GAN, have been introduced. Rumi-GANs leverage negative examples to enhance the generator's ability to learn positive examples. We evaluate the performance of two controllable GAN variants, CGAN and Rumi-GAN, in generating game levels targeting specific constraints of interest: playability and controllability. This evaluation is conducted under two scenarios: with and without the inclusion of negative examples. The goal is to determine whether incorporating negative examples helps the GAN models avoid generating undesirable outputs. Our findings highlight the strengths and weaknesses of each method in enforcing the generation of specific conditions when generating outputs based on given positive and negative examples.

\end{abstract}
%==================================================
\section{Introduction}

% An overview of the project. What has been done in the area, what problem are you trying to solve, what is your approach, and what did you find?
% state of the world ...
In traditional generative models without conditioning, there is limited control over the attributes of the generated data. This limitation is effectively addressed by conditional models, where additional information can condition the data generation process towards specific outcomes.
% current limitations ...
Recent research by \citet{asokan2020teaching} introduces a method where GAN models are trained using both positive and negative examples to guide and avoid specific outcomes (Rumi-GAN). Positive examples are data samples that align well with the desired outcome or target distribution. These examples represent the ideal or goal that the GAN is trying to achieve.  Negative examples, on the other hand, are data samples that do not meet the desired outcome or diverge from the target distribution.

% therefore we did ...
This study investigates the performance of two controllable GAN generators (CGAN and Rumi-GAN) through three sets of experiments conducted in the context of two 2D tile-based games. In each experiment, our goal is to enforce the generation of specific constraints within each game level. The first experiment aims to generate playable levels, emphasizing playability as the targeted constraint. The second experiment builds on the first by controlling a game-related feature, such as the specific number of pipes or treasures within the levels, while maintaining the playability constraint.

The first controllable model, a Conditional GAN (CGAN), is trained with levels containing both desired and undesired conditions (positive and negative examples), incorporating additional label information about the level conditions into the data. The second model, Rumi-GAN, also receives levels with both desired and undesired conditions and employs a loss function to encourage the generation of level segments with desired conditions while discouraging those with undesired conditions. Additionally, we utilize a baseline control model, a vanilla GAN, which cannot incorporate negative examples and, therefore, only receives levels with the desired condition for training.

% what we found ...
Our results show that incorporating negative examples in training GAN models can help enforce some constraints of interest like playability.
% our contributions are ...
To our knowledge, this study is the first attempt to systematically compare various controllable GAN models and evaluate their effectiveness in conditioning generated outputs. Additionally, the integration of negative examples in GAN training (Rumi-GAN) has previously been only utilized on state-of-the-art computer vision datasets like MNIST, CelebA, and CIFAR-10, and this work is the first to apply this new approach to game-level segments.

The codebase of the project (including the training database, trained models, and generated artifacts) is available on GitHub\footnote{\url{https://github.com/MahsaBazzaz/NegativeExamples}}

%==================================================
\section{Related Work}
% What other work has been done in the area? Cite at least 5 other sources in this section.

A range of studies have explored the use of GAN models for game level generation. \citet{Kumaran2020} developed a GAN-based architecture capable of generating levels for multiple distinct games from a common gameplay action sequence. \citet{volz2018evolving} trained a GAN to generate Super Mario Bros levels, using a variety of fitness functions to guide the search for levels with specific properties. \citet{Giacomello2018} applied GANs to learn a model of DOOM levels, finding that the inclusion of topological features improved the similarity of generated levels to human-designed ones. \citet{capps2021using} focused on generating whole Mega Man levels by using multiple GANs to model levels with different types of segments. \citet{schrum2020interactive} used interactive exploration of the GAN latent space to generate Mario levels and dungeons. \citet{awiszus2020toad} proposed TOAD-GAN that expands SinGAN architecture, enabling it to create token-based levels while being trained on a single example.

Conditional GANs have also been utilized in image generation and more recently in game level generation.  
\citet{torrado2020bootstrapping} proposed a new GAN architecture, CESAGAN, for video game level generation, which incorporated an embedding feature vector input to condition the training of the discriminator and generator. This approach also introduced a bootstrapping mechanism to reduce the number of levels necessary for training and generate a larger number of playable levels with fewer duplicates.  \citet{Kelvin2020} applied conditional generative adversarial networks (CGANs) to create road maps, providing a balance between user control and automatic generation. \citet{hald2020procedural} furthered the application of GANs by using parameterized GANs to produce levels for a puzzle game, although they encountered challenges in approximating certain conditions.

Outside the domain of game level generation, \citet{asokan2020teaching} introduces a novel method for training GANs called ``Rumi-GAN''. This approach incorporates the concept of negative samples, which the GAN learns to avoid. The Rumi-GAN enhances the discriminator's capability to accurately model the target distribution and expedites the learning process of the generator. This method has been validated through various experiments, demonstrating its effectiveness, especially in scenarios involving imbalanced datasets where certain classes are under-represented.

Inspired by this approach, we aim to apply negative examples to the generation of game levels. Specifically, we intend to compare the performance of vanilla GAN, CGAN, and Rumi-GAN when provided with both positive and negative examples of game levels.

%==================================================
\section{Domains}
% What training data did you use, where did you find it, what did you need to do to get it usable, etc? Refer to the class resources if needed.
This work uses the Sturgeon \citep{cooper2022sturgeon}
constraint-based level generator to create a corpus of Super Mario Bros \cite{GAME_mariobros} level segments. These segments are $14\times32$ in size and are based on the level 1-1 from the VGLC \citep{summerville_vglc_2016}. We have also created a corpus of a custom game called \texttt{Cave} which is a $14\times14$ simple top-down cave map and it was first introduced by \citet{cooper2022constraint} using tiles from Kenney \cite{WEB_kenney}.

In this work, we used two of Sturgeon's constraint-based features to generate our corpus. First is the ability to add constraints on the number of specific tiles. Second is (in addition to generating playable levels) the ability to create unplayable levels using an unreachability constraint \citep{cooper2024literally} that are similar to playable levels in local tile patterns, but are not possible to play.

We created $3000$ each of playable and unplayable Mario segments with exactly $1$,$2$, and $3$ pipes in them, resulting in $9000$ playable segments and $9000$ unplayable segments of Mario. We also created $3000$ each of playable and unplayable Cave segments with exactly $1$, $2$, and $3$ treasures in them, resulting in also $9000$ playable segments and $9000$ unplayable segments of Cave. 

Table \ref{tab:data_representation} shows different tile types and corresponding symbols in each game. The ``start'' and ``end'' tiles are specifically kept in each level as the minimum requirement for level playability. We use one-hot encoding of these level segments during GAN training.

\begin{table}[bt]
    \begin{minipage}[b]{0.6\linewidth}
        \centering
    \begin{tabular}{|>{\centering\arraybackslash}p{0.45\linewidth}|>{\centering\arraybackslash}p{0.2\linewidth}|}
        \hline
        \multicolumn{2}{|c|}{Mario} \\
        \hline
        Type & Symbol \\
        \hline
        Ground & X \\ 
        \hline
        Breakable & S \\ 
        \hline
        Empty & - \\ 
        \hline 
        Question Block & Q \\ 
        \hline
        Top Left Pipe & \textless \\ 
        \hline
        Top Right Pipe & \textgreater  \\ 
        \hline
        Left Pipe & [ \\ 
        \hline
        Right Pipe & ] \\ 
        \hline
        Start & \{ \\ 
        \hline
        End & \} \\ 
        \hline
    \end{tabular}
    \end{minipage}\hfill
    \begin{minipage}[b]{0.4\linewidth}
        \centering
        \begin{tabular}{|>{\centering\arraybackslash}p{0.45\linewidth}|>{\centering\arraybackslash}p{0.3\linewidth}|}
            \hline
            \multicolumn{2}{|c|}{Cave} \\
            \hline
            Type & Symbol \\
            \hline
            Solid & X \\ 
            \hline
            Empty & - \\ 
            \hline 
            Treasure & 2 \\ 
            \hline 
            Start & \{ \\ 
            \hline
            End & \} \\ 
            \hline
        \end{tabular}
    \end{minipage}
    \caption{Different tile types and corresponding symbols present in each game.}
    \label{tab:data_representation}
\end{table}

%==================================================
\section{System Overview}
All three controllable GANs adopt the Deep Convolutional GAN architecture as used by \citet{volz2018evolving}, which is based on the original work of \citet{pmlr-v70-arjovsky17a}. This architecture employs batch normalization in both the generator and discriminator after each layer, ReLU activation functions for all layers in the generator (instead of Tanh), and LeakyReLU activation in all layers of the discriminator.

These models are trained using the WGAN algorithm. Wasserstein Generative Adversarial Networks offer an alternative to traditional GAN training, providing more stable training as demonstrated by \citet{pmlr-v70-arjovsky17a}. Both the generator and discriminator are trained with RMSprop, using a batch size of 32 and a default learning rate of 0.00005 for $200$ iterations. 
% Initially we intended to train all models for the same number of iterations but the slow training rate of CGAN compared to Vanilla and Rumi GANs, required this model to need almost 3 times more training hours to achieve the same number of iterations as other models. Therefore, we decided to train all models for the same number of hours, rather than the same number of iterations, to ensure a fair comparison of their performance. We will report the training rate of these models in each experiment as well.

The following sections detail each controllable model, describing the classes of data used for training and the objective function of the training process.

\subsection{Vanilla Generative Adversarial Nets}
The vanilla GANs as introduced by \citep{goodfellow2014generative}, consist of two models: a generative model (G) that captures the data distribution, and a discriminative model (D) that estimates the probability of a sample being real (from the training data) or fake (from the generative model). The entire training process is framed as a minimax game, where the discriminator tries to maximize the objective function, as described in Equation \ref{eq:vanilla} \citep{goodfellow2014generative}, while the generator tries to minimize it. Here, $x \sim p_d$ represents samples from the real data distribution (positive examples), and $x \sim p_g$ represents samples from the generator's distribution.

\begin{equation}
    \min_{p_g} \max_{D(\boldsymbol{x})} \left( \mathbb{E}_{\boldsymbol{x} \sim p_d}[\log D(\boldsymbol{x})] + \mathbb{E}_{\boldsymbol{x} \sim p_g}[\log (1 - D(\boldsymbol{x}))] \right)
    \label{eq:vanilla}
\end{equation}

In this model, controllability is achieved by training a separate model for each set of desired constraints. For instance, with the goal of generating Mario levels with a specific number of pipes, a vanilla GAN is trained specifically to generate Mario level segments containing only one pipe. This means that the model is trained exclusively on level segments that feature only a single pipe.

\subsection{Conditional Generative Adversarial Networks}
CGANs~\citep{mirza2014conditional} apply extra information $y$ to both the discriminator and generator. As shown in Equation \ref{eq:cgan} \citep{mirza2014conditional} The loss function of a CGAN is an extension of the vanilla GAN loss function, incorporating this conditional information. This again results in in a minimax game in which the generator tries to generate realistic data conditioned on $y$, while the discriminator tries to distinguish between real and generated data, also conditioned on $y$. Here 
$x \sim p_{data}$ includes both positive and negative samples, distinguished by label $y$.

\begin{equation}
\begin{aligned}
\min_{G} \max_{D} V(D, G) = & \ \mathbb{E}_{\boldsymbol{x} \sim p_{\text{data}}(\boldsymbol{x})}[\log D(\boldsymbol{x} \mid \boldsymbol{y})] \\
& + \mathbb{E}_{\boldsymbol{z} \sim p_{z}(\boldsymbol{z})}[\log (1 - D(G(\boldsymbol{z} \mid \boldsymbol{y})))]
\end{aligned}
\label{eq:cgan}
\end{equation}

Since positive and negative examples vary depending on the training target, a specific CGAN must be trained for each desired condition in each experiment. The only difference between these models is the label of the examples, which changes based on the model's objective.

\subsection{Rumi Generative Adversarial Nets}
Rumi-GAN \citep{asokan2020teaching} is a specialized type of GAN inspired by the Sufi poet Rumi's philosophy of learning from both positive and negative experiences. Equation \ref{eq:rumi} \citep{asokan2020teaching} shows how in this approach data distribution $p_d$ is split into the target distribution that the GAN is required to learn (positive samples, $p_d^+$) and the distribution of samples that it must avoid (negative samples, $p_d^-$). The fake distribution which are the samples drawn from the generator $p_{g}$ is there as before. $\alpha^{+}$ and $\alpha^{-}$ is a weighting factor for the positive and negative real data distribution term. We set $\alpha^{+}$ to $1$ and $\alpha^{-}$ to $0.5$.

\begin{equation}
\begin{aligned}
\mathcal{L}_{D}^{S}=-\Bigl(&\alpha^{+} \mathbb{E}_{\boldsymbol{x} \sim p_{d}^{+}}[\log D(\boldsymbol{x})]+\mathbb{E}_{\boldsymbol{x} \sim p_{g}}[\log (1-D(\boldsymbol{x}))] \\
& +\alpha^{-} \mathbb{E}_{\boldsymbol{x} \sim p_{d}^{-}}[\log (1-D(\boldsymbol{x}))]\Bigr)
\end{aligned}
\label{eq:rumi}
\end{equation}

Again, since positive and negative examples vary depending on the training target, we need to train a specific Rumi-GAN for each desired constraint.

%==================================================
\section{Experiments}
\renewcommand{\arraystretch}{1.2}
\begin{table*}[h]
    \begin{tabularx}{\textwidth}{|l|l|l|X|}
    \hline
    & \textbf{Model} & \textbf{Goal} & \textbf{Input Data}\\
    \hline
    \multirow{5}{*}{\rotatebox[origin=c]{90}{Experiment 1}}
    & Vanilla GAN& playable segments & playable segments of class 1, 2, and 3 \\
    \cline{2-4}
    & Rumi-GAN & playable  segments &  playable segments of class 1, 2, and 3 (+)\\
    & & & unplayable segments of class 1, 2, and 3 (-)\\ 
    \cline{2-4}
    & CGAN & playable segments & playable segments of class 1, 2, and 3 with labels (1,0), (2,0), (3,0) \\
    & & & unplayable segments of class 1, 2, and 3 with labels (1,1), (2,1), (3,1) \\ 
    
    \hline
    % \multirow{9}{*}{\rotatebox[origin=c]{90}{Experiment 3}}
    % & Rumi & generation of class 1 & playable segments of class 1 (+), class 2 (-), class 3 (-) \\
    % & Rumi & generation of class 2 & playable segments of class 1 (-), class 2 (+), class 3 (-)\\
    % & Rumi & generation of class 3 & playable segments of class 1 (-), class 2 (-), class 3 (+)\\ 
    % & Conditional & generation of class 1 &  playable segments of class 1 label (1,0) \\ 
    % & & & playable segments of class 2, and 3 with labels (2,1) and (3,1) \\
    % & Conditional & generation of class 2 &  playable segments of class 2 label (2,0) \\ 
    % & & & playable segments of class 1, and 3 with labels (1,0) and (3,0) \\
    % & Conditional & generation of class 3 &  playable segments of class 3 label (3,0) \\ 
    % & & & playable segments of class 1, and 2 with labels (1,0) and (2,0) \\
    % \hline
    
    \multirow{18}{*}{\rotatebox[origin=c]{90}{Experiment 2}}
    & \multirow{2}{*}{Vanilla GAN} & generation of class 1 & playable segments of class 1 \\
    \cline{3-4}
    & & generation of class 2 & playable segments of class 2 \\ 
    \cline{3-4}
    & & generation of class 3 & playable segments of class 3 \\ 
    \cline{2-4}
    & \multirow{2}{*}{Rumi-GAN} & generation of class 1 & playable segments of class 1 (+), class 2 (-), class 3 (-)\\
    & & & unplayable segments of class 1, 2, and 3 (-) \\
    \cline{3-4}
    &  & generation of class 2 & playable segments of class 1 (-), class 2 (+), class 3 (-)\\
    & & & unplayable segments of class 1, 2, and 3 (-) \\
    \cline{3-4}
    &  & generation of class 3 & playable segments of class 1 (-), class 2 (-), class 3 (+)\\ 
    & & & unplayable segments of class 1, 2, and 3 (-) \\
    \cline{2-4}
    & \multirow{2}{*}{CGAN} & generation of class 1 &  playable segments of class 1 label (1,0) \\ 
    & & & playable segments of class 2, and 3 with labels (2,1) and (3,1) \\
    & & & unplayable segments of class 1, 2, and 3 with labels (1,1), (2,1), and (3,1) \\
    \cline{3-4}
    &  & generation of class 2 &  playable segments of class 2 label (2,0) \\ 
    & & & playable segments of class 1, and 3 with labels (1,1) and (3,1) \\
    & & & unplayable segments of class 1, 2, and 3 with labels  (1,1), (2,1), and (3,1)\\
    \cline{3-4}
    &  & generation of class 3 &  playable segments of class 3 label (3,0)\\ 
    & & & playable segments of class 1, and 2 with labels (1,1) and (2,1) \\
    & & & unplayable segments of class 1, 2, and 3 with labels (1,1), (2,1), and (3,1) \\
    \hline
    \end{tabularx}
    \caption{Input data and goals of trained models in experiment 1. The positive and negative signs indicate the positive and negative examples in Rumi-GAN.}
    \label{tab:experiment_settings}
\end{table*}

\begin{figure}[t]
\centering
\includegraphics[width=0.47\textwidth]{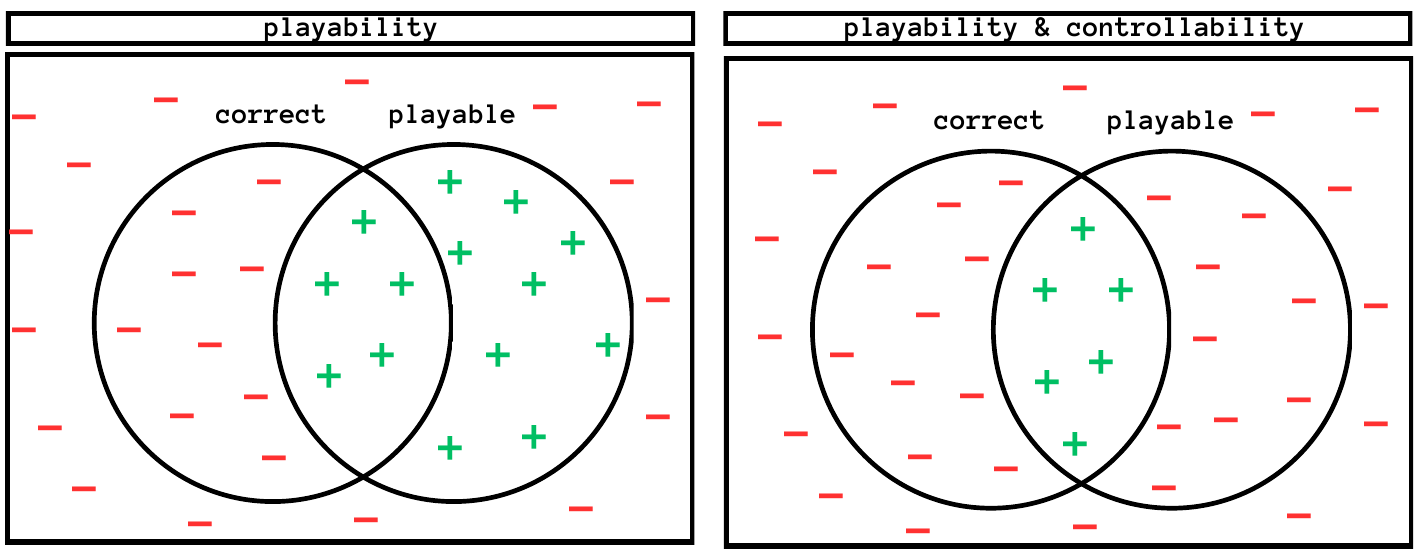}
\caption{Venn diagrams of the sample space. Green plus signs represent positive samples and red minus signs represent negative samples provided to models. }
\label{fig:first}
\end{figure}

We conducted two different experiments to examine two sets of constraints we would like to see in the generated levels: playability and controllability. A playable level is a level such that there exists a path between the level's start and end locations. Controllability is derived from the correctness of the number of controlled features (eg. pipes or treasures) at each level. A level is considered correct if it has the desired number of features. Figure \ref{fig:first} shows the Venn diagram of the sample space. The sample space is divided into \texttt{playable-correct}, \texttt{playable-incorrect}, \texttt{unplayable-correct}, and \texttt{unplayable-incorrect} subspaces. Positive and negative samples are chosen from these subspaces according to the model's objective. In the experiments, negative examples are added in addition to the positive data that models utilize. To ensure an unbiased distribution of training data for models that incorporate negative examples, an equal number of positive and negative examples are sampled for training. Table \ref{tab:experiment_settings} the models trained in each experiment and the exact input data of each model. 

\subsection{Experiment One: Playability}
In this experiment, the goal is to ensure playability as the primary constraint for the models to meet. This means that we take the positive samples($p_d^+$) from the \texttt{playable} subspace and negative samples ($p_d^-$) from the \texttt{unplayable} subspace. Both Conditional GAN and Rumi-GAN use this negative distribution as additional information.
 
% \subsection{Experiment Three \todo{swap}}
% In this experiment, the goal is to enforce the number of some features as a condition for the models to satisfy. Specifically, we focus on the number of pipes in Mario levels and the number of treasures in Cave levels. This approach gets positive samples($p_d^+$) from the \texttt{correct} subspace and the negative samples($p_d^-$) from the\texttt{incorrect} subspace, which Conditional GAN and Rumi-GAN use this negative distribution as additional information. 

\subsection{Experiment Two: Playability and Controllability}
This experiment's goal is to enforce both the number of some features in the level (number of pipes in Mario and number of treasures in Cave) and playability as a constraint for the models to satisfy. This approach gets positive samples ($p_d^+$) from the \texttt{playable-correct} subspace and negative samples($p_d^-$) from the \texttt{playable-incorrect},  \texttt{unplayable-correct}, and \texttt{unplayable-incorrect} subspaces. This means Rumi-GAN, and Conditional GAN get the playable samples with the desired condition as the positive examples to learn, and playable samples without the condition (other playable classes) and all unplayable classes as negative examples. 

%==================================================
\section{Evaluation}
After the training models in each experiment, we generated $500$ levels with each trained model and we evaluated each model based on the criteria of that experiment.

\subsection{Experiment One: Playability}
 To evaluate Experiment One, we measure the percentage of playable levels as the metric. We use Sturgeon to find (if available) the shortest path between the start and goal of the level segments. Generated level segments that don't have a start or end, or have multiple starts and ends count as unplayable levels immediately. Table \ref{tab:experiment1} shows the results of this experiment. In Mario both models using negative examples show better performance than vanilla GAN. In Cave, only Rumi-GAN takes advantage of the negative examples resulting in better performance compared to the other models.

\begin{table}[h!]
  \centering
  \tablefontsize % Apply the custom font size
\tablefontsize % Apply the custom font size
    \begin{tabularx}{0.47\textwidth}{ cccc }
    \hline
    & \textbf{Vanilla GAN} & \textbf{Rumi-GAN} & \textbf{CGAN} \\
    \hline
    Mario  & 67.8\% & 72\% &  \textbf{75.4\%} \\
    \hline
    Cave  & 87\% & \textbf{89.6\%} & 66.6\% \\
    \hline
\end{tabularx}
\caption{Experiment One Results. Percentage of playable results.}
  \label{tab:experiment1}
\end{table}
% \subsection{Experiment Three}
% In the evaluation of experiment three, we measure the percentage of levels that achieved the goal of the model which is levels with the correct number of controlled pipes/treasures. Note that the objective is only focused on the controllability and unplayable levels with a correct number of controlled features count as success in this experiment. Table \ref{tab:experiment3} shows the results of this experiment.

% \begin{table}[h!]
%   \centering
%   \tablefontsize % Apply the custom font size
%     \begin{tabularx}{0.47\textwidth}{ |X|X|X|X| }
%     \hline
%     & Vanilla & Rumi & Conditional \\
%     \hline
%     Mario  &  &  &  x\\
%     \hline
%     Cave  &  &  & x\\
%     \hline
% \end{tabularx}
%   \caption{Experiment Three Results. Percentage of correct outputs.}
%  \label{tab:experiment3}
% \end{table}

\subsection{Experiment Two: Playability and Controllability}
To evaluate Experiment Two, we measure the percentage of levels that have the correct number of pipes/treasures while being playable. This means unplayable levels with the correct number of pipes/treasures, or playable levels with an incorrect number of pipes/treasures count as failures in this experiment as they only achieved half of the objective. Note that for the purposes of evaluating the number of treasures in Cave, it is only the presence of the treasures that matters, not whether they are reachable. Tables \ref{tab:exp2mario} and \ref{tab:exp2cave} present the results of this experiment. Incorporating both the playability constraint and the number of pipes/treasures, as expected,  makes generating correct outputs more challenging than in Experiment One. This increased difficulty leads to fewer \texttt{playable} levels being produced by each model compared to Experiment One. As a result, there are also fewer \texttt{playable-correct} outputs. Overall, while the inclusion of negative examples slightly improved performance in CGAN for Mario levels, it did not provide any benefit for Cave levels.

% \begin{table}[h!]
%     \centering
%     \tablefontsize
%     \begin{tabularx}{0.47\textwidth}{ |X|X|X|X| }
%     \hline
%     & Vanilla GAN & Rumi-GAN & CGAN \\
%     \hline
%     % \multirow{2}{*}{Mario}
%      Mario & 24\% & 18.8\% & \textbf{25\%}\\
%      % & 66\% playable  & 65\% playable  &  65\% playable \\
%      % & 35\% correct &  28\% correct &  37\% correct \\
%     \hline
%     % \multirow{2}{*}{Cave}
%      Cave & \textbf{13.6\%} & 13.1\% & 12.8\%\\
%      % & 82\% playable & 83\% playable &  23\% playable \\
%      % & 16\% correct & 15\% correct &  22\% correct \\
%     \hline
% \end{tabularx}
% \caption{Experiment Two Results. Aggregated percentage of playable and correct outputs.}
% \label{tab:experiment2}
% \end{table}

\begin{table*}[h!]
\centering
\begin{adjustbox}{width=0.7\textwidth}
\begin{tabular}{lcccccccccccc}
\toprule
& \multicolumn{4}{c}{\textbf{correct}} & \multicolumn{4}{c}{\textbf{playable}} & \multicolumn{4}{c}{\textbf{playable correct}} \\
\cmidrule(lr){2-5} \cmidrule(lr){6-9} \cmidrule(lr){10-13}
& 1 & 2 & 3 & Avg & 1 & 2 & 3 & Avg & 1 & 2 & 3 & Avg \\
\midrule
Vanilla & 44.8 & 41.2 & 19.0 & 35.0 & 67.6 & 65.0 & 64.4 & 65.6 & 30.4 & 28.8 & 12.8 & 24.0 \\
Rumi & 41.2 & 41.0 & 0.6 & 27.6 & 65.2 & 71.2 & 56.8 & 64.4 & 25.4 & 31.0 & 0.0 & 18.8 \\
Conditional & 44.4 & 31.2 & 34.8 & \textbf{36.6} & 65.4 & 69.8 & 62.4 & \textbf{65.8} & 29.4 & 22.4 & 23.4 & \textbf{25.0} \\
\bottomrule
\end{tabular}
\end{adjustbox}
\caption{Experiment Two Results. Detailed percentage of playable and correct outputs of Mario}
\label{tab:exp2mario}
\end{table*}

\begin{table*}[h!]
\centering
\begin{adjustbox}{width=0.7\textwidth}
\begin{tabular}{lcccccccccccc}
\toprule
& \multicolumn{4}{c}{\textbf{correct}} & \multicolumn{4}{c}{\textbf{playable}} & \multicolumn{4}{c}{\textbf{playable correct}} \\
\cmidrule(lr){2-5} \cmidrule(lr){6-9} \cmidrule(lr){10-13}
& 1 & 2 & 3 & Avg & 1 & 2 & 3 & Avg & 1 & 2 & 3 & Avg \\
\midrule
Vanilla & 24.4 & 20.6 & 4.6 & 16.5 & 85.0 & 79.0 & 81.8 & 81.9 & 20.6 & 16.2 & 4.0 & \textbf{13.6} \\
Rumi & 18.4 & 24.6 & 1.8 & 14.9 & 83.6 & 84.0 & 82.2 & \textbf{83.2} & 16.2 & 21.4 & 1.8 & 13.1 \\
Conditional & 38.6 & 30.0 & 19.6 & \textbf{29.4} & 41.7 & 31.9 & 32.2 & 35.3 & 16.0 & 9.6 & 12.8 & 12.8 \\
\bottomrule
\end{tabular}
\end{adjustbox}
\caption{Experiment Two Results. Detailed percentage of playable and correct outputs of Cave}
\label{tab:exp2cave}
\end{table*}

\begin{figure*}[!h]
\centering
\includegraphics[width=0.8\textwidth]{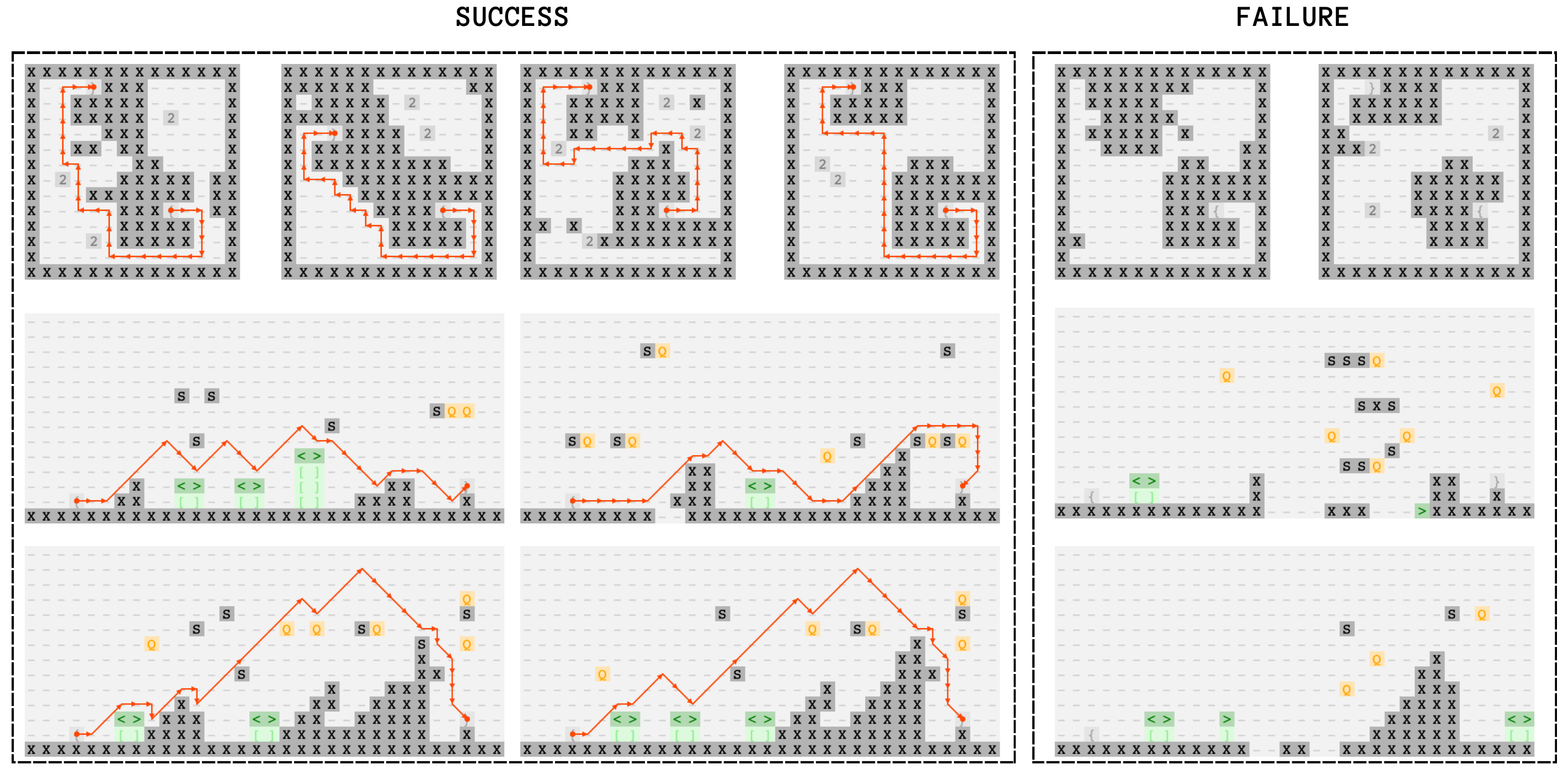}
\caption{Examples of successful and failed levels generated in Experiment 1. A successful level is defined as a playable level. In contrast, unplayable levels either lack start or end positions or have blocked player paths.}
\label{fig:artifacts1}
\end{figure*}

\begin{figure*}[!h]
\centering
\includegraphics[width=0.8\textwidth]{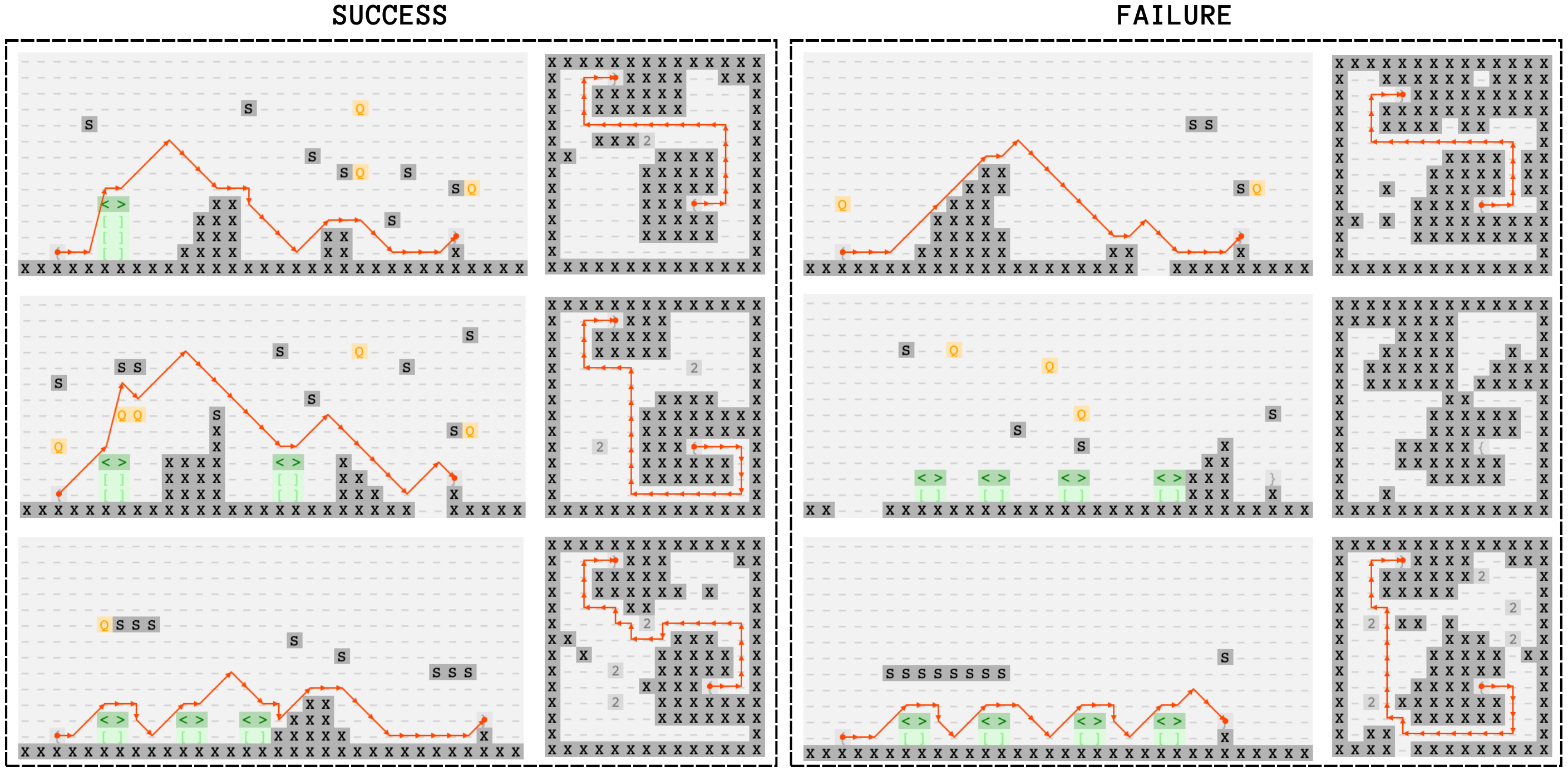}
\caption{Examples of successful and failed levels generated in Experiment 2. A successful level must have the exact correct features (pipes/treasures) and be playable. Failed levels may be playable but with an incorrect number of features, or unplayable but with the correct number of features.}
\label{fig:artifacts2}
\end{figure*}
It's important to note that the results of Experiment Two reinforce the findings from Experiment One, with the same models—CGAN and RUMI-GAN—showing superior performance in enforcing playability constraints. Specifically, CGAN emerged as the leading model in terms of constraint correctness. However, the results did not indicate a clear winning model for enforcing both playability and correctness constraints. We believe this suggests that when combining different constraints, the model must be able to distinguish between the negative examples of each constraint. Otherwise, as shown in Experiment Two, the model may not derive significant benefits from the combination. 

%==================================================
\section{Discussion}

The results of Experiment One demonstrate that incorporating negative examples in training GAN models can enhance their ability to generate more playable levels. In future work, we aim to reinforce this approach using levels annotated with players' paths.  The surprising decrease in the performance of GAN models using negative examples in Experiment Two, may suggest the importance of the quality of the negative examples. In future work, we would like to explore multi-stage training approaches with high-quality negative examples in fine-tuning steps. We believe this approach could be effective when combined with bootstrapping methods \citep{torrado2020bootstrapping}, or active learning methods with minimal training levels \citep{bazzaz2023active}. These approaches could make the training easier, with the only price for the additional controllability being the number of models trained on the minimal data.

%==================================================
\section{Conclusion}

% Briefly summarize the overall project. What could be done in the future to build on the work?
This study explores the potential advantages of integrating negative examples into Generative Adversarial Networks (GANs) to enhance the generation of game levels. Inspired by the work of \citet{asokan2020teaching} on Rumi-GAN, the primary focus lies in leveraging positive examples to guide GANs toward producing desired outputs and negative examples to avoid undesirable outputs. Through comparative analyses involving Conditional GANs (CGANs), Rumi-GANs, and a baseline vanilla GAN with and without negative examples, it was observed that incorporating negative examples improves the capability of models to generate more playable outputs. However, this enhancement does not necessarily aid in enforcing constraints related to the controllability of specific features, such as the number of features in game levels.

\section{Acknowledgments}
Support provided by Research Computing at Northeastern University (\url{https://rc.northeastern.edu/}).

\bigskip

\bibliography{refs}

\end{document}